\documentclass[11pt]{article}

\usepackage[preprint]{acl}
\usepackage{makecell}
\usepackage{graphicx}
 \usepackage{microtype}
 \usepackage{amsmath} 
 \usepackage{float}
 \usepackage{multicol}
\usepackage{afterpage}
\usepackage[table]{xcolor}
\usepackage{booktabs}
\usepackage{tablefootnote}
\usepackage{multirow}  
\usepackage{soul}
\usepackage{times}
\usepackage{latexsym}
\usepackage[most]{tcolorbox}
\usepackage{url}
\usepackage{makecell}
\usepackage{bbm}
\usepackage{amsmath}
\usepackage{arydshln}
\usepackage{multirow}
\usepackage{booktabs}
\usepackage{amssymb}
\usepackage{pifont}
\usepackage{float}
\usepackage{listings}
\usepackage{tabularx}
\usepackage{footnote}
\makesavenoteenv{table}
\makesavenoteenv{table*}
\makesavenoteenv{tabular}
\usepackage[ruled,vlined,linesnumbered]{algorithm2e}
\usepackage{fontawesome5}
\newtcolorbox{prompt}[1]{
    enhanced,
    drop shadow=black!5!white,
    left=4mm,
    right=4mm,
    top=1mm,
    bottom=1mm,
    boxsep=0mm,
    rounded corners,
    title=#1,
    fontupper=\scriptsize\linespread{1}\fontfamily{lmr}\selectfont,
    }
\usepackage[english,bidi=default]{babel} 
\babelfont{rm}{TeXGyreTermesX} 

\title{DeepVision-103K: A Visually Diverse, Broad-Coverage, and Verifiable Mathematical Dataset for Multimodal Reasoning}

\author{
 \textbf{Haoxiang Sun\textsuperscript{1,2}},
 \textbf{Lizhen Xu\textsuperscript{1,2}},
\textbf{Bing Zhao\textsuperscript{1}},
\textbf{Wotao Yin\textsuperscript{1}},
\\
 \textbf{Wei Wang\textsuperscript{1}},
 \textbf{Boyu Yang\textsuperscript{1}},
 \textbf{Rui Wang\textsuperscript{2}\footnotemark[2]},
 \textbf{Hu Wei\textsuperscript{1}}\footnotemark[2]
\\
 \textsuperscript{1}Alibaba Group,
 \textsuperscript{2}Shanghai Jiao Tong University
\\
\\
  \faGithub~\href{https://github.com/SKYLENAGE-AI/DeepVision-103K}{\texttt{https://github.com/SKYLENAGE-AI/DeepVision-103K}}
    \\
   \href{https://huggingface.co/datasets/skylenage/DeepVision-103K}{%
  \includegraphics[height=0.8em]{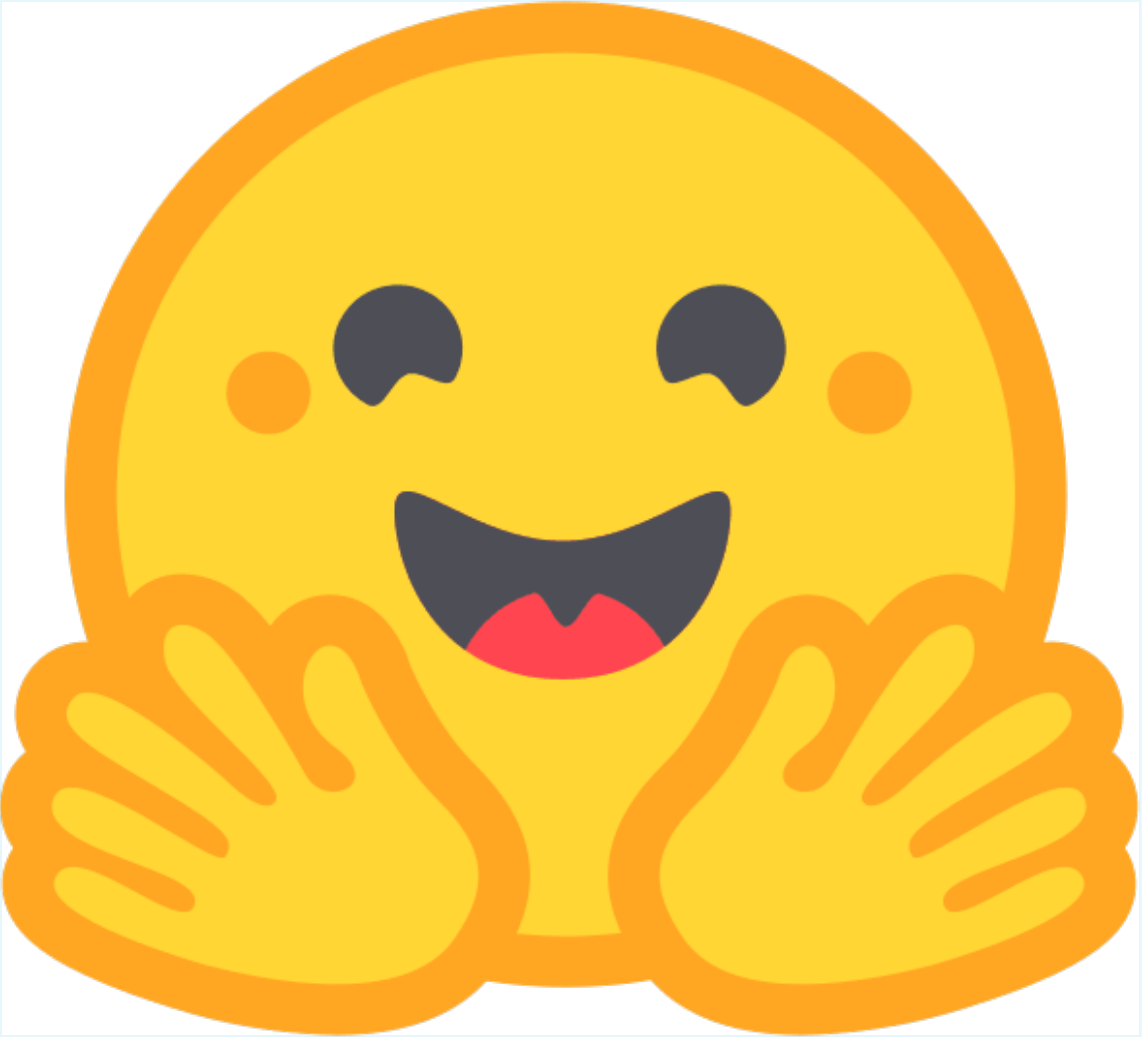}\,
  \texttt{https://hf.co/datasets/skylenage/DeepVision-103K}}
 }

\begin{document}

\maketitle


\begin{figure*}[!h]
    \centering
    \includegraphics[width=\textwidth]{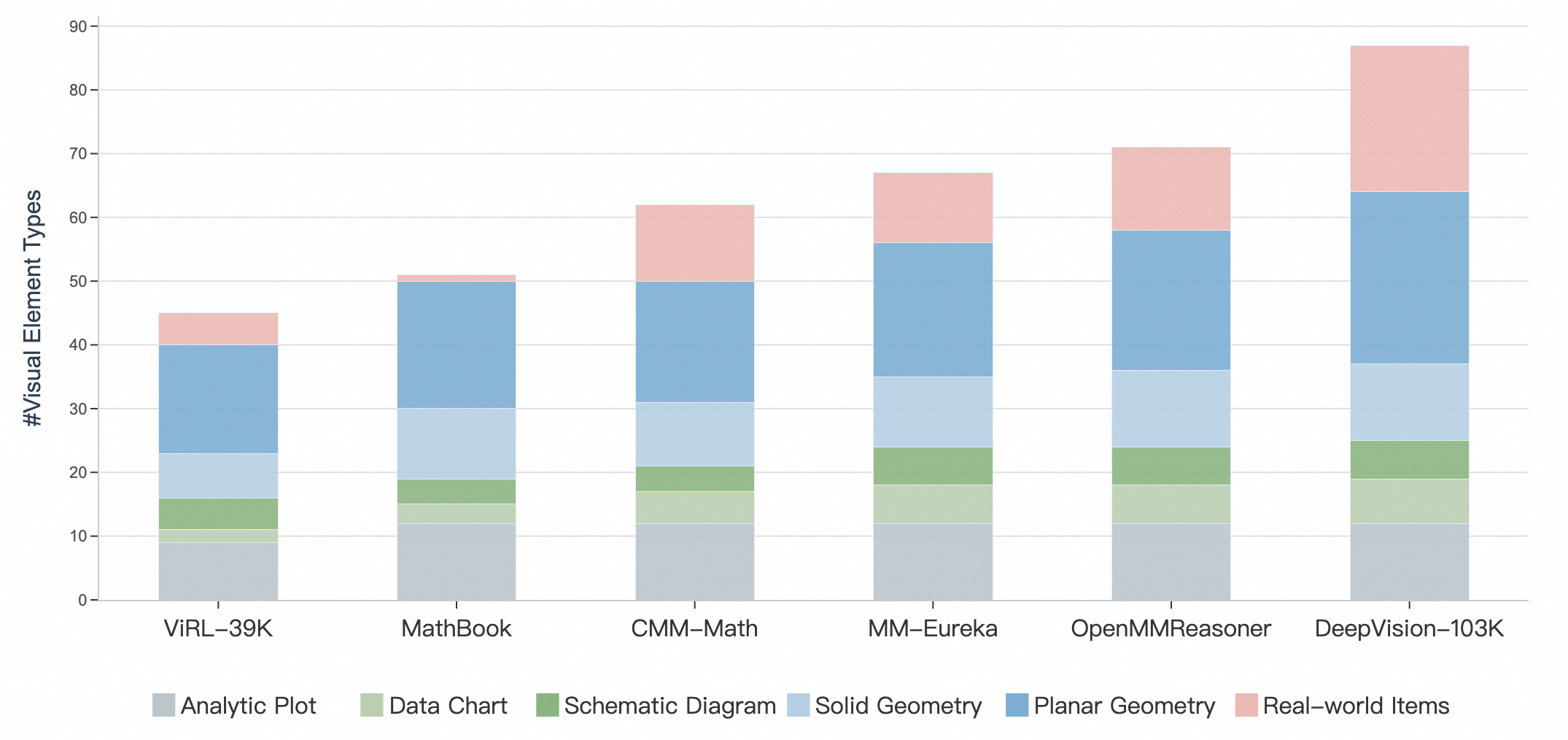}
    \caption{The number of different visual element types of training datasets.}
    \label{fig:visual}
\end{figure*}

\begin{figure*}[!h]
    \centering
    \includegraphics[width=\textwidth]{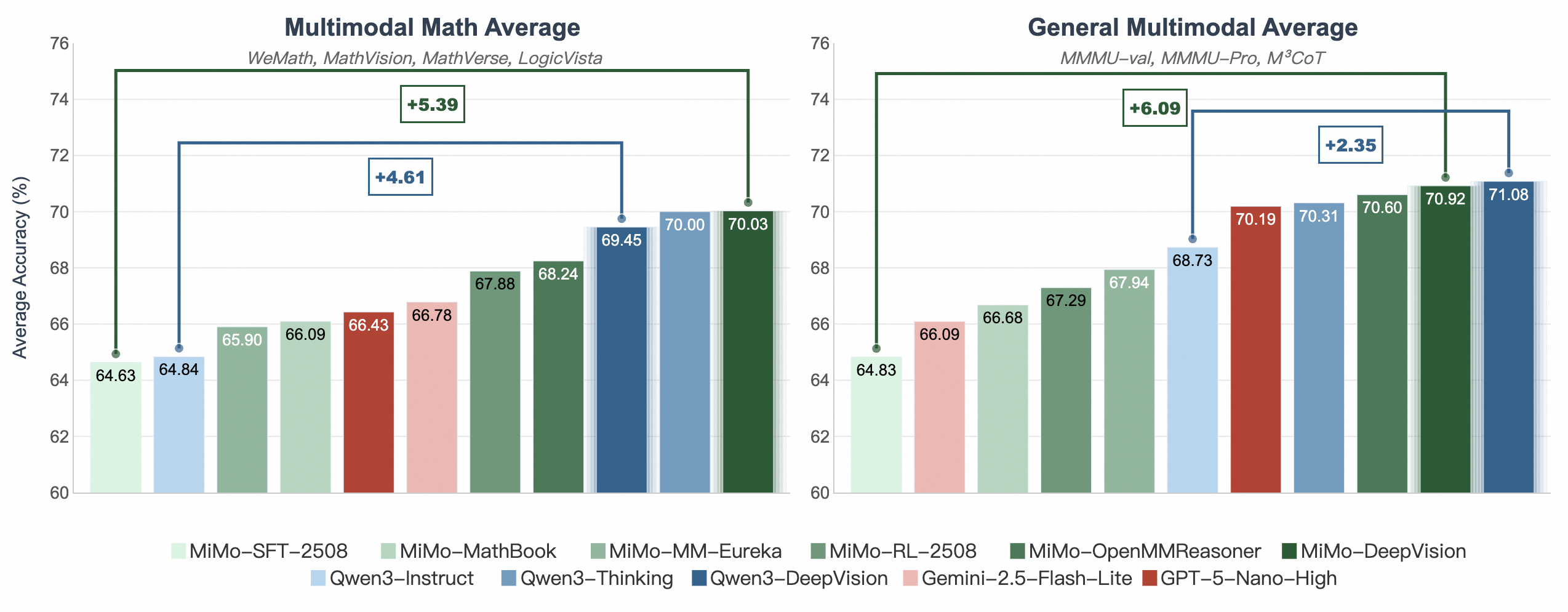}
    \caption{Performance on multimodal math and general multimodal benchmarks, we report averaged Pass@1 accuracy across benchmarks.}
    \label{fig:perf}
\end{figure*}

\begin{abstract}
Reinforcement Learning with Verifiable Rewards (RLVR)  has been shown effective in enhancing the visual reflection and reasoning capabilities of Large Multimodal Models (LMMs). However, existing datasets are predominantly derived from either small-scale manual construction or recombination of prior resources,  which limits data diversity and coverage, thereby constraining further gains in model performance. To this end, we introduce \textbf{DeepVision-103K}, a comprehensive dataset for RLVR training that covers diverse K12 mathematical topics, extensive knowledge points, and rich visual elements. Models trained on DeepVision achieve strong performance on multimodal mathematical benchmarks, and generalize effectively to general multimodal reasoning tasks. Further analysis reveals enhanced visual perception, reflection and reasoning capabilities in trained models, validating DeepVision's effectiveness for advancing multimodal reasoning.
\end{abstract}

\vspace{1mm}
\section{Introduction}

Large language models (LLMs) trained with reinforcement learning from verifiable rewards (RLVR), such as DeepSeek-R1~\cite{deepseek-r1} and OpenAI o-series~\cite{openai-o1}, demonstrate remarkable reasoning capabilities. A key insight is that RLVR incentivizes thinking behaviors—the ability to decompose problems, self-correct in step-by-step reasoning. 
Recent works ~\cite{vl-rethinker,visionaryr1,visionthink} extend this paradigm to large multimodal models (LMMs), achieving enhanced visual reflection and reasoning abilities. Central to this progress is high-quality training data, but existing training sets for multimodal RLVR exhibit several key limitation.

\begin{itemize}
\item Synthetically constructed datasets: Fully synthesized with professional tools like GeoGebra~\cite{geo3k,wemath20}. They provide abundant data for constructible categories (e.g., geometric diagrams, function curves) but \textbf{lack real-world mathematical scenarios}, limiting robust generalization to general tasks.

\item Human-annotated K12 datasets: Gathered from authentic K12 education scenarios and human-annotated to obtain verifiable answers ~\cite{mmeureka,CMM-Math}. While offering broader categories, \textbf{dependence on expert annotation limits its scalability}.

\item Recombination of existing datasets: Filtration~\cite{mcts,visiong1} or recombination~\cite{lmmr1,r1onevision,openmmreasoner} of prior sources. These approaches create no novel problems, resulting in overlap across datasets and \textbf{lacking broader data distribution}.
\end{itemize}

To address these limitations, we propose \textbf{DeepVision-103K}, a large-scale multimodal mathematical dataset designed for RLVR, featuring:

\begin{itemize}

    \item \textbf{Visual Diversity}: DeepVision-103K covers major visual categories including geometry, analytic plots, charts, and real-world items in mathematical contexts. Within each category, DeepVision offers richer element types than existing open-source datasets (Figure \ref{fig:visual}).
    
    \item \textbf{Broad Coverage}: DeepVision-103K incorporates wide-ranging multimodal mathematical problems (Figure \ref{fig:domain}) and visual logic problems (mazes, chess, tetris), jointly enhancing mathematical and visual logic reasoning.

    \item \textbf{Automatic Data Curation Pipeline}: We present an automatic curation pipeline (Figure \ref{fig:curation}) comprising validity filtering, pass-rate stratification and correctness verification, which transforms diverse but noisy real-world K12 problems into structured and verifiable QA pairs.
\end{itemize}

Consequently, models trained on DeepVision-103K achieve top performance (Figure~\ref{fig:perf}) on mathematical and general multimodal reasoning. DeepVison models outperform: (1) models trained on other open-source datasets, (2) the official thinking variant built on the same base model, and (3) strong closed-source baselines. These results underscore the value of DeepVision-103K as a resource for advancing multimodal reasoning. The remainder of this paper is organized as follows:
\begin{itemize}
    \setlength{\itemsep}{0em}
    \item Sec.~\ref{sec:overview} presents an overview of DeepVision-103K, including its format, visual elements distribution, and topics covered.
    \item Sec.~\ref{sec:data_curation} details the data curation pipeline to construct DeepVision-103K, encompassing validity filtering, model-centric difficulty filtering and query correctness verification.
    \item Sec.~\ref{sec:exp} describes the training setup and evaluation results of models trained on DeepVision-103K.
    \item Sec.~\ref{sec:ana} explores how training on DeepVision-103K enhances model capabilities and presents ablation studies of the data curation pipeline.
\end{itemize}

\vspace{-2mm}
\section{Overview of DeepVision-103K}
\label{sec:overview}

\begin{figure}[!h]
    \centering
    \vspace{-4mm}
    \includegraphics[width=0.48\textwidth]{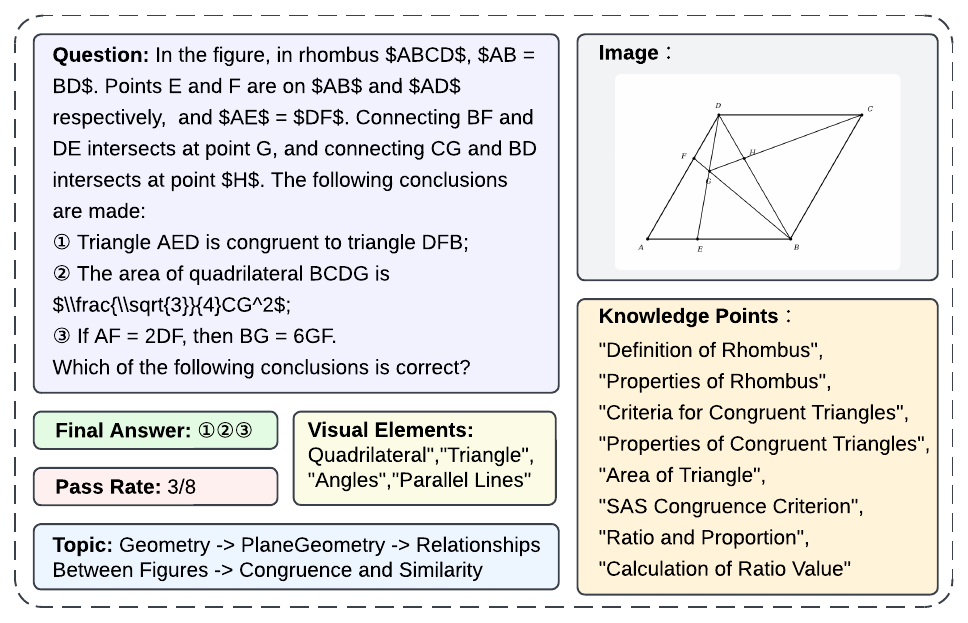}
    \caption{A data sample from DeepVision-103K.}
    \vspace{-2mm}
    \label{fig:overview}
\end{figure}
\vspace{-2mm}
\noindent DeepVision-103K adopts a rich annotation schema to facilitate various downstream tasks in multimodal reasoning. As illustrated in Figure~\ref{fig:overview}, each sample contains the following components:
\begin{table}[!h]
\centering
\resizebox{\linewidth}{!}{
\begin{tabular}{p{0.22\linewidth} p{0.74\linewidth}}
\hline
\textbf{Field} & \textbf{Description} \\
\hline
\textbf{Question \& Image} & A multimodal mathematical problem consisting of a textual problem statement and the corresponding image. \\
\hline
\parbox[t]{0.22\linewidth}{\textbf{Final}\\\textbf{Answer}} & A unique, verifiable answer that enables rule-based reward computation in RLVR. \\
\hline
\textbf{Pass Rate} & The proportion of correct responses obtained during model rollouts. \\
\hline
\textbf{Topic} & A hierarchical classification indicating which branch of mathematics the problem belongs to. \\
\hline
\textbf{Knowledge Points} & A list of specific mathematical concepts, theorems, or techniques required to solve the problem. \\
\hline
\parbox[t]{0.22\linewidth}{\textbf{Visual}\\\textbf{Elements}} & A list of geometric or graphical objects depicted in the image, describing what visual content should be perceived and interpreted. \\
\hline
\end{tabular}
}
\caption{Annotation fields and definitions.}
\label{tab:annotation_fields}
\end{table}

\vspace{-1mm}
\subsection{Visual Diversity}

\begin{figure}[!h]
    \centering
    \vspace{-4mm}
    \includegraphics[width=0.48\textwidth]{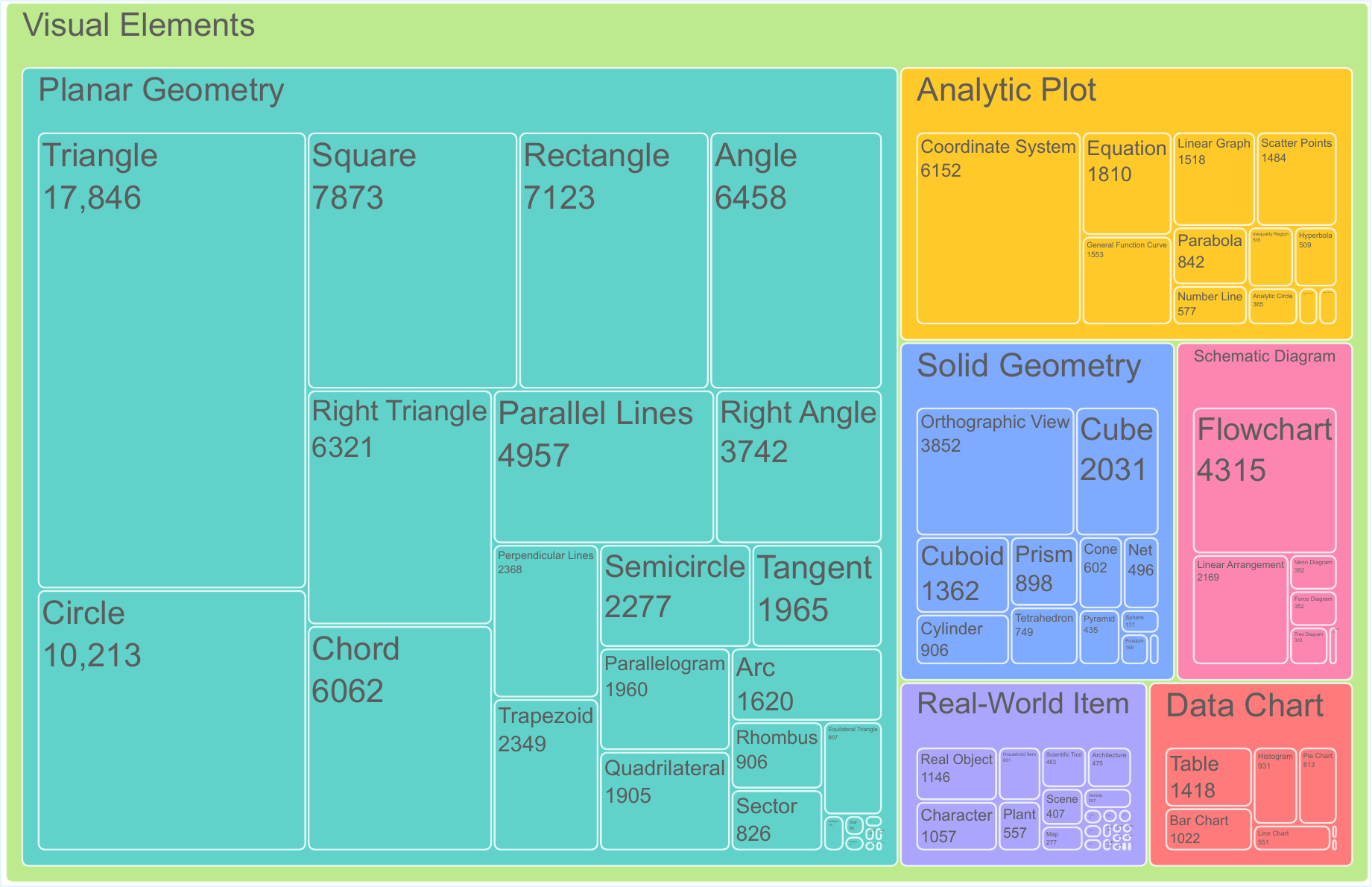}
    \caption{Visual elements in DeepVision-103K.}
    \label{fig:visual elements}
    \vspace{-1mm}
\end{figure}

To assess the richness of visual content in DeepVision, we built a taxonomy based on~\cite{a,b} then instructed GPT-5 mini to  annotate the \emph{visual elements} in each image with both \emph{categories} and \emph{fine-grained types}. Prompts and other implementation details are provided in Appendix \ref{visual-anno}. DeepVision includes diverse visual elements across 6 categories (Figure~\ref{fig:visual elements}), each presenting unique perceptual challenges.

We summarized  the coverage of each category in Table \ref{tab:visual_categories}. Notably, DeepVision captures \textbf{cross-category visual combinations} and real-world items in mathematical contexts, requiring models to reason across multiple visual representations simultaneously. Examples are provides in Appendix~\ref{appendix:visual_examples}.

\begin{table}[!h]
\centering
\resizebox{\columnwidth}{!}{%
\begin{tabular}{@{}ll@{}}
\toprule
\textbf{Category} & \textbf{Key Visual Elements} \\
\midrule
\texttt{Planar Geometry} & \makecell[l]{Primitives (Angle, Triangle, Circle, Quadrilateral,Polygon), \\ Relations (Parallelism, Tangency, Chords), \\ Properties (Right Angles, Perpendicularity)} \\
\midrule
\texttt{Solid Geometry} & \makecell[l]{3D Primitives (Cube, Prism, Cylinder, Cone), \\ Spatial Representations (Orthographic Views, Nets), \\ Sections (Frustums, Hemispheres)} \\
\midrule
\texttt{Analytic Plot} & \makecell[l]{Coordinate Systems, Function Curves (Linear, General), \\ Conic Sections (Parabola, Hyperbola), \\ Scatter Points, Inequality Regions} \\
\midrule
\texttt{Data Chart} & \makecell[l]{Statistical Graphs (Bar, Histogram, Pie, Line), \\ Structured Data (Tables, Stem-and-Leaf)} \\
\midrule
\texttt{Schematic Diagram} & \makecell[l]{Logical Structures (Flowcharts, Tree Diagrams), \\ Physics/Sets (Force Diagrams, Circuits, Venn Diagrams), \\ Linear Arrangements} \\
\midrule
\texttt{Real-World Item} & \makecell[l]{Objects (Characters, Household Items), \\ Contextual Scenes (Architecture, Maps, Scientific Tools)} \\
\midrule
\texttt{Cross-category} & Combinations of multiple visual categories \\
\bottomrule
\end{tabular}%
}
\caption{Visual categories and element coverage in DeepVision-103K.}
\vspace{-4mm}
\label{tab:visual_categories}
\end{table}

\subsection{Broad Coverage}
DeepVision-103K covers a broad range of mathematical topics and knowledge points. We categorized each problem using a hierarchical topic structure following  ~\citet{wemath20}.

\begin{figure}[!h]
    \centering
    \includegraphics[width=0.45\textwidth]{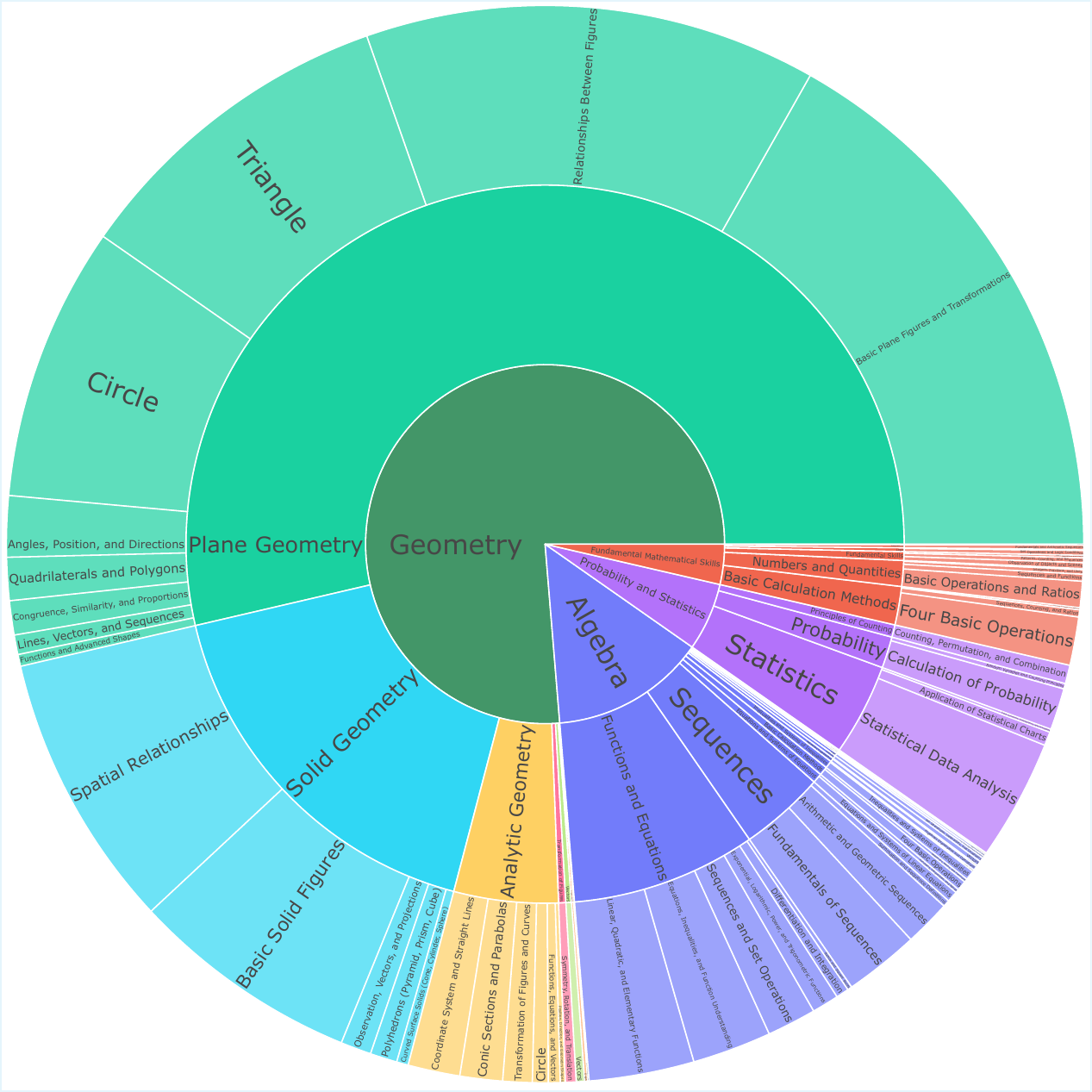}
    \caption{Mathematical topics in DeepVision-103K.}
    \label{fig:domain}
\end{figure}

As shown in Figure~\ref{fig:domain}, our dataset spans four major mathematical disciplines. \textbf{Geometry} accounts for the largest share, followed by substantial coverage of \textbf{Algebra}, \textbf{Probability and Statistics}, and \textbf{Fundamental Mathematical Skills}. Across these domains, DeepVision includes over 200 fine-grained topics and nearly 400 distinct knowledge points, exposing models to diverse problem-solving patterns and fostering more robust, generalizable reasoning. Beyond formula- and theorem-based mathematics, DeepVision also incorporates \textbf{visual logic problems} from Zebra-CoT~\cite{zebracot} and GameQA \cite{gamerl}—including maze, chess, tetris, games where solutions emerge primarily from visual perception and logical deduction.

\section{Construction of DeepVision-103K}
\label{sec:data_curation}

\begin{figure}[!h]
    \centering
    \includegraphics[width=0.48\textwidth]{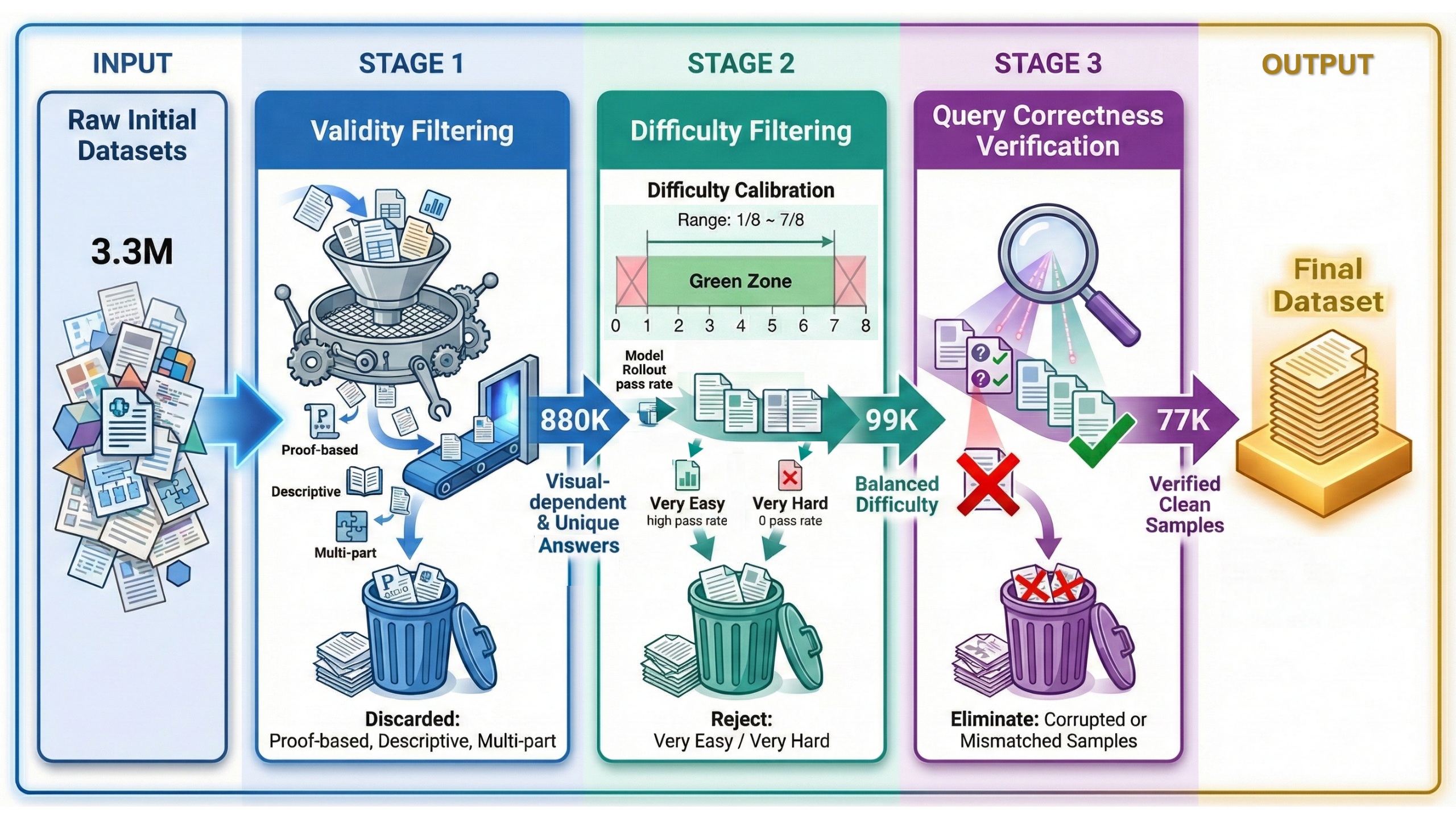}
    \caption{Curation pipeline for mathematical data in  DeepVision-103K.}
    \label{fig:curation}
\end{figure}

\noindent We curated our dataset from open-source multimodal mathematics SFT corpora, including MM-MathInstruct-3M~\citep{mmmath} and MultiMath-300K~\citep{multimath}. Both datasets collect K12 level problems from real educational contexts, forming an initial pool of 3.3M samples. To derive verifiable data from this extensive yet noisy collection, we applied a three-stage curation pipeline in Figure \ref{fig:curation}:

\begin{enumerate}
    \item \textbf{Validity Filtering:} Remove problems inherently unsuitable for RL training, including proof-based, descriptive and multi-answer questions.
    
    \item \textbf{Difficulty Filtering:} Calibrate sample difficulty based on model capability through rollout pass rates.
    
    \item \textbf{Query Correctness Verification:} Validate the correctness of image-question pairs and answers to eliminate corrupted samples.
\end{enumerate}

\paragraph{Stage 1: Validity Filtering.}
Reinforcement learning requires unique and verifiable answers to provide reliable reward signals. In this stage, we first applied rule-based filtering to remove proof or explanation tasks containing keywords such as ``prove'', ``explain'', ``describe''. For the remaining questions, we employed Qwen3-VL-32B-Instruct~\citep{qwen3vl} to analyze each sample, counting the number of answers and determining whether visual information is necessary. Only questions with unique answer and genuinely require visual information were retained. After this stage, we obtained 880K questions.

\paragraph{Stage 2: Difficulty Filtering.} Data with appropriate difficulty is crucial for efficient RL training \cite{nlbj}. 
DeepMath~\citep{deepmath} employed SOTA models to annotate difficulty based on human-defined standards, which may not align well with model capabilities~\cite{wemath20}. We adopted an approach similar to Qwen3-VL~\citep{qwen3vl}. 
For each question, we performed 8 rollouts using MiMo-VL-7B-SFT~\citep{mimo} and then calculated accuracy with MathVerify~\cite{mathverify}. 
We keep samples whose pass rate falls in $[\tfrac{1}{8}, \tfrac{7}{8}]$. Zero-pass samples are discarded as they are either too hard or unverifiable, while full-pass samples are removed because overly easy data can reduce exploration during RL training \cite{rlzoo}. For visual-logic data, which is well-formed from Zebra-CoT~\cite{zebracot}, GameQA \cite{gamerl} and other sources, we apply the same rollout-and-filtering pipeline and obtain 26K clean, verifiable training examples. Appendix~\ref{appendix:Difficulty Filtering} provides further details.

\paragraph{Stage 3: Query Correctness Verification.}
Correct answers are essential for reliable RL rewards, and so are well-formed questions. Although we filtered out zero-pass samples, models still randomly guessed answers for inherently problematic queries (e.g., garbled text or image-text mismatches). To this end, we prompted Gemini-3-Flash~\citep{gemini3} to (1) verify that each question is complete and free of corrupted text, (2) detect potential image–text mismatches, and (3) validate the provided answer. We retained only samples that pass all three checks. Details of the verification protocol are provided in Appendix~\ref{appendix:Correctness Verification}. After this final stage, we obtained 77K correct and verifiable QA pairs for RL training.

\begin{table*}[!t]
\centering
\resizebox{\textwidth}{!}{
\begin{tabular}{@{}lccccccc@{}}
\toprule
\multirow{2}{*}{\textbf{Model}} & \multicolumn{4}{c}{\textbf{Multimodal Math}} & \multicolumn{3}{c}{\textbf{General Multimodal}} \\
\cmidrule(lr){2-5} \cmidrule(lr){6-8}
& \textbf{WeMath} & \textbf{MathVision} & $\textbf{MathVerse}_{\textbf{vision}}$ & \textbf{LogicVista} & $\textbf{MMMU}_{\textbf{val}}$ & $\textbf{MMMU}_{\textbf{Pro}}$ & $\textbf{M}^{3}\textbf{CoT}$ \\
\midrule
\multicolumn{8}{l}{\textbf{\textit{Closed-source Models}}} \\
GPT-5-Nano-High & 78.62 & 58.75 & 70.30 & 58.03 & 70.78 & 70.64 & 69.15 \\
Gemini-2.5-Flash-Lite & 83.85 & 52.47 & 70.30 & 60.49 & 64.77 & 65.08 & 68.42 \\
\midrule
 \multicolumn{8}{l}{\textbf{\textit{Qwen3-VL-8B Series}}} \\
Qwen3-VL-8B-Instruct & 79.36 & 51.44 & 67.38 & 61.16 & 67.66 & 67.69 & 70.83 \\
Qwen3-VL-8B-Thinking & 84.54 & \textbf{57.89} & \textbf{72.84} & 64.73 & 69.33 & \textbf{70.29} & 71.31 \\
\rowcolor{blue!5} Qwen3-VL-8B-DeepVision & \textbf{85.11} & 55.49 & 72.46 & \textbf{64.73} & \textbf{71.33} & \textbf{70.29} & \textbf{71.61} \\
\midrule
\multicolumn{8}{l}{\textbf{\textit{MiMo-VL-7B Series}}} \\
MiMo-VL-7B-SFT-2508 & 74.42& 50.69 & 72.71 & 60.71 & 63.77 & 60.69 & 70.02 \\
MiMo-VL-7B-RL-2508 & 76.95 & 53.91 & \textbf{76.39} & 64.28 & 67.44 & 63.87 & 70.57 \\
MiMo-VL-7B-MM-Eureka & 79.08 & 50.00 & 73.35 & 61.16 & 67.67 & 65.78 & 70.36 \\
MiMo-VL-7B-MathBook & 77.18 & 51.31 & 73.60 & 62.28 & 66.33 & 63.47  & 70.23 \\
MiMo-VL-7B-OpenMMReasoner & \textbf{83.45} & 52.97 & 74.87 & 61.68 & 66.78 & 66.82 & \textbf{78.21}\footnote{Extremely high because OpenMMReasoner includes ViRL-39K\cite{virl}, which includes $\text{M}^{3}\text{CoT}$. }\\
\rowcolor{blue!5} MiMo-VL-7B-DeepVision & 82.98 & \textbf{55.24} & 76.26 & \textbf{65.62} & \textbf{71.00} & \textbf{69.19} & 72.56 \\
\bottomrule
\end{tabular}
}
\caption{Performance comparison across multimodal mathematical reasoning and general multimodal benchmarks. We report Pass@1 accuracy (\%). The best results for each model family are shown in \textbf{bold}.}
\label{tab:main_results}
\end{table*}

\section{Experiments}
\label{sec:exp}
In this section, we present a comprehensive evaluation of the mathematical and general multimodal reasoning capabilities of models trained on DeepVision.

\subsection{Setup}

\paragraph{Models}
We conducted training on LMMs that already possess thinking capabilities, including MiMo-VL-7B-SFT-2508 \cite{mimo} and Qwen3-VL-8B-Instruct \cite{qwen3vl}. Both models have been exposed to visual reasoning data during the pretrain or midtrain stages, exhibiting native visual thinking abilities.

\paragraph{Algorithm} We employed GSPO~\citep{gspo} for RL training, utilizing rule-based rewards based on answer correctness (+1 for correct answers, 0 otherwise). We specified the required response format through prompts, and no additional format reward was applied. Detailed training configurations and prompts are provided in Appendix~\ref{app:training setting}.

\paragraph{Baselines} We compared against (1) \textbf{Closed-source models}: GPT-5-Nano-High, Gemini-2.5-Flash-Lite; (2) \textbf{Official thinking variants}: Qwen3-VL-8B-Thinking, MiMo-VL-7B-RL-2508; and (3) \textbf{Open-source datasets}: MM-Eureka~\citep{mmeureka}, human-annotated real K12 data; MathBook~\citep{wemath20}, human curated data; OpenMMReasoner~\citep{openmmreasoner}, filtration and combination of  prior sources. We trained MiMo-VL-7B-SFT-2508 on these datasets under the same setting for fair comparison with MiMo-VL-7B-DeepVision.

\paragraph{Evaluation}
We evaluated our models on the following benchmarks: (1) \textbf{Multimodal Math}: WeMath~\citep{bench:wemath}, $\text{MathVerse}_{\text{vision}}$~\citep{bench:mathverse}, MathVision~\citep{bench:mathvision}, and LogicVista~\citep{bench:logicvsita}.
(2) \textbf{General Multimodal}:  $\text{MMMU}_{\text{VAL}}$~\citep{bench:mmmu}, $\text{MMMU}_{\text{Pro\_full}}$~\cite{mmmu_pro} and $\text{M}^{3}{\text{CoT}}$~\citep{m3cot}.
For inference parameters, we set the maximum token length at 32K for all evaluation. Decoding parameters follow the official recommendations. Complete details are provided in Appendix~\ref{app:eval setting}.

\subsection{Multimodal Mathematics Reasoning Results}

As shown in Table~\ref{tab:main_results}, training on DeepVision yields strong results in mathematical reasoning.

\paragraph{Consistent gains across benchmarks.} Compared to respective Instruct/SFT baselines, Qwen3-VL-8B-DeepVision and MiMo-VL-7B-DeepVision achieve uniform improvements across all evaluated benchmarks, with gains ranging from 2.91\% to 8.56\%.

\paragraph{Substantial improvements.} On WeMath and LogicVista, DeepVision models surpass their official thinking variants and closed-source models. Qwen3-VL-8B-DeepVision reaches sota results on WeMath (85.11\%), MiMo-VL-7B-DeepVision reaches sota results on LogicVista (65.62\%). On MathVision and MathVerse, they exceed or substantially narrow the gap with thinking variants.

\paragraph{Superiority over existing open-source datasets.} Compared to models trained on other open-source datasets, MiMo-VL-7B-DeepVision demonstrates clear advantages, highlighting the value of DeepVision as a high-quality RL training resource.

\subsection{Generalization Beyond Mathematics}
Table~\ref{tab:main_results} shows that
DeepVision models generalize effectively to general-purpose multimodal tasks, achieving consistent improvements over foundation models and surpassing official thinking variants across all three benchmarks. In contrast, models trained on other open-source datasets show limited improvements in general domains. This disparity suggests that the diverse visual elements and broad domain coverage in DeepVision are crucial for enhancing general multimodal reasoning capabilities, which is further supported by our analysis in Sec.~\ref{visual-reasoning}.

\section{Analyses}
\label{sec:ana}
Our analyses investigate the following key questions:

\noindent\textit{\textbf{Q1}: Enhanced Capabilities.} What capabilities are enhanced after RL on DeepVision-103K? 

\noindent\textit{\textbf{Q2}: The Value of Visual Logic Data.} What role do the introduced visual logic tasks (e.g., mazes, tangrams, and games) play in the DeepVision-103K dataset? 

\noindent\textit{\textbf{Q3}: Necessity of query correctness verification.} Recent studies \cite{rom,spuriousreward} suggest that RLVR can work even under random rewards. Is correctness verification step truly necessary in our data curation pipeline?

\subsection{Enhanced Capabilities}
Training on DeepVision-103K presents increasing response length, upward rewards and stable entropy (Appendix \ref{training curves}). 
To further investigate how RL on DeepVision improves model capabilities, we systematically compared Qwen3-VL-8B-Instruct and Qwen3-VL-8B-DeepVision across multiple benchmarks. We collected cases where DeepVision succeeds but Instruct fails and asked human annotators to analyze the underlying mechanism following Algorithm \ref{alg:annotation}.

\begin{algorithm}[!h]
\caption{Human Annotation Protocol}
\label{alg:annotation}
\SetAlgoLined
\DontPrintSemicolon
\KwIn{Query $(\text{Image}, \text{Text})$,Ground Truth $y$, Incorrect Instruct Response \textcolor{red}{$R_{I}$}, Correct DeepVision Response \textcolor{green!60!black}{$R_{D}$}}
\KwOut{Improvement Mechanism $C$}

Analyze visual descriptions in \textcolor{red}{$R_{I}$}\;
\eIf{Descriptions contradict $\text{Image}$}{
    Root Cause $\leftarrow$ \textbf{Visual Misperception}\;
}{
    Root Cause $\leftarrow$ \textbf{Incorrect Reasoning}\;
}

\uIf{Root Cause is \textbf{Visual Misperception}}{
    \eIf{\textcolor{green!60!black}{$R_{D}$} correct at first observation}{
        $C \leftarrow$ \textsc{Visual Perception}\;
    }{
        \eIf{\textcolor{green!60!black}{$R_{D}$} corrected via reflection}{
            $C \leftarrow$ \textsc{Visual Reflection}\;
        }{
            $C \leftarrow$ \textsc{Guess}\;
        }
    }
}
\ElseIf{Root Cause is \textbf{Incorrect Reasoning}}{
    \eIf{\textcolor{green!60!black}{$R_{D}$} shows valid reasoning chain}{
        $C \leftarrow$ \textsc{Reasoning}\;
    }{
        $C \leftarrow$ \textsc{Guess}\;
    }
}
\Return $C$\;
\end{algorithm}

\begin{figure}[!h]
    \centering
    \includegraphics[width=0.48\textwidth]{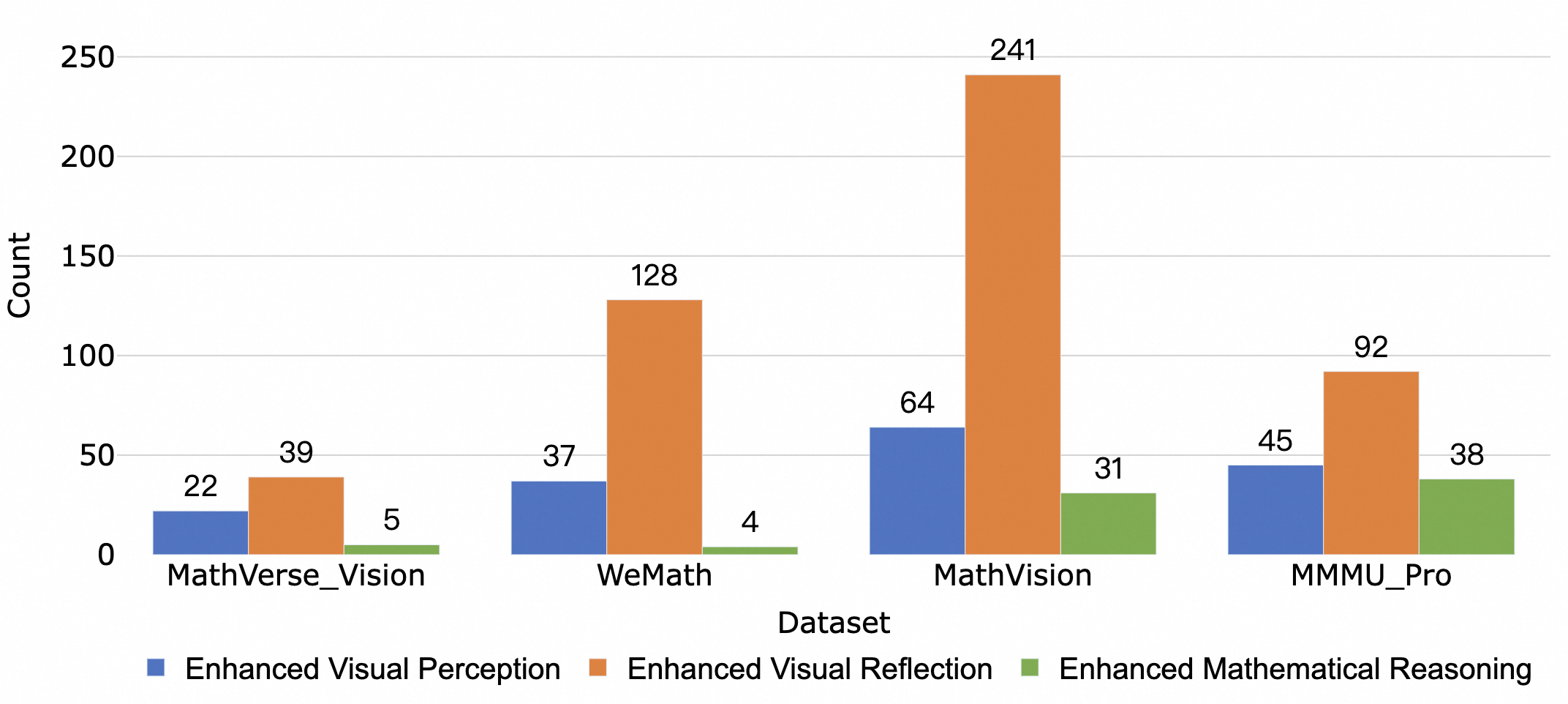}
    \caption{Enhanced Capabilities}
    \label{fig:enhancement_distribution}
\end{figure}

\begin{figure}[!h]
    \centering
    \includegraphics[width=0.48\textwidth]{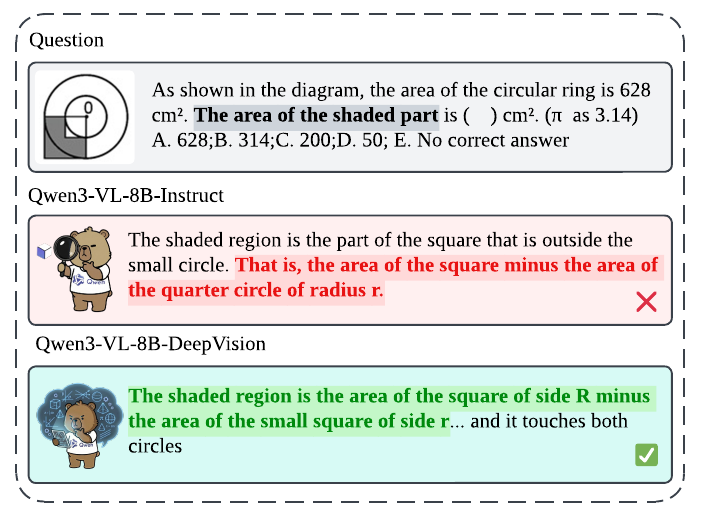}
    \caption{DeepVision model correctly identifies the shaded region on the first attempt.}
    \label{fig:better_perception}
\end{figure}

For each sample, annotators cited verbatim evidence from model response (Figure \ref{fig:better_perception}). If no evidence supports, the sample was labeled as \textsc{Guess}. Our analysis reveals three enhancement types, as shown in Figure~\ref{fig:enhancement_distribution}.

\vspace{-2mm}
\paragraph{Type I: Enhanced Visual Perception.}
We observed enhanced ``one-shot perception''—DeepVision model correctly identifies geometric shapes, numerical values, and spatial relationships in the initial observation, without requiring iterative re-examination (Figure \ref{fig:better_perception}).

\begin{figure}[!h]
    \centering
    \includegraphics[width=0.48\textwidth]{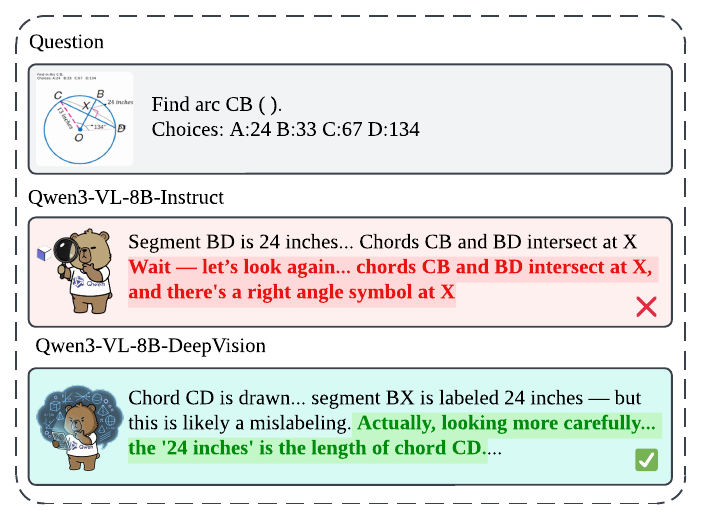}
    \caption{DeepVision model actively re-examines visual content to correct errors, while the base model merely rephrases without genuine verification.}
    \label{fig:better_reflection}
\end{figure}

\paragraph{Type II: Enhanced Visual Reflection.}
When initial perceptual errors occur, DeepVision demonstrates a stronger capacity for genuine visual re-examination—actively recounting elements, remeasuring angles, and re-inspecting spatial relationships—whereas the base model tends to rephrase conclusions without revisiting the visual content (Figure \ref{fig:better_reflection}).

\begin{figure}[!h]
    \centering
    \includegraphics[width=0.48\textwidth]{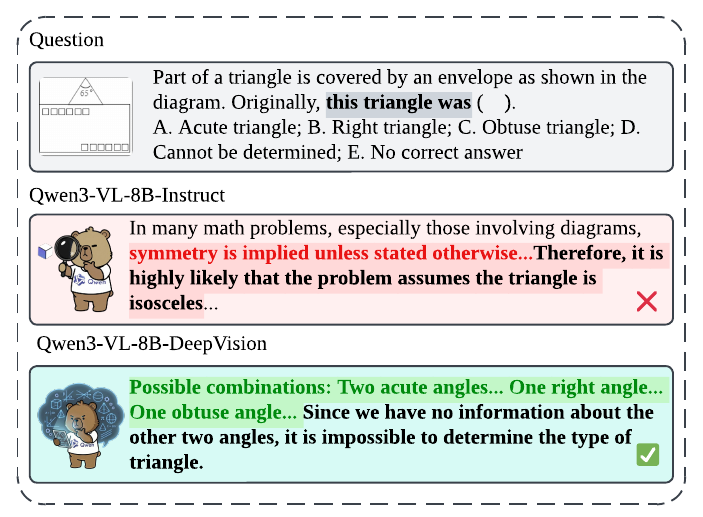}
    \caption{DeepVision model systematically enumerates all possible angle combinations and concludes the type cannot be determined, while the Instruct model incorrectly assumes symmetry without justification.}
    \label{fig:better_reasoning}
\end{figure}

\paragraph{Type III: Enhanced Mathematical Reasoning.}
Beyond visual capabilities, RL fine-tuning also enhances pure mathematical reasoning. In cases where both models correctly extract identical visual information, DeepVision demonstrates more rigorous mathematical reasoning (Figure \ref{fig:better_reasoning}).

\subsection{The Value of Visual Logic Data}
\label{visual-reasoning}

\begin{table*}[!h]
\centering
\resizebox{\linewidth}{!}{
\begin{tabular}{lccccc|cccc}
\toprule
\multirow{2}{*}{\textbf{Data Composition}} & \multicolumn{5}{c}{\textbf{Multimodal Math}} & \multicolumn{4}{c}{\textbf{General Multimodal}} \\
& WeMath & MathVision & MathVerse & LogicVista & Avg. & MMMU$_\text{val}$ & MMMU$_\text{pro}$ & M3CoT & Avg. \\
\midrule
MiMo-VL-7B & 74.42 & 50.69 & 72.71 & 60.71 & 64.63 & 63.77 & 60.69 & 70.02 & 64.83 \\
DeepVision-103K$_{200}$ & \textbf{82.98} & 55.23 & \textbf{76.26} & \textbf{65.92} & \textbf{70.10} & \textbf{71.00} & 69.19 & 72.56 & \textbf{70.92} \\
\midrule
\multicolumn{10}{l}{\textit{w/o visual logic data}} \\
\quad Math-77K$_{150}$ & 81.67 & 54.83 & 74.23 & 63.98 & 68.68 & 70.00 & 68.55 & 72.09 & 70.21 \\
\quad Math-77K$_{200}$ & 82.07 & \textbf{55.72} & 74.74 & 63.53 & 69.02 & 68.50 & \textbf{69.67} & \textbf{72.65} & 70.27 \\
\midrule
\multicolumn{10}{l}{\textit{w/o multimodal math data}} \\
\quad Visual-logic-26K$_{50}$ & 79.54 & 51.61 & 73.35 & 63.98 & 67.12 & 68.33 & 67.34 & 71.61 & 69.09 \\
\midrule
\multicolumn{10}{l}{\textit{w/o correctness verification}} \\
\quad Unverified-125K$_{200}$ & 82.36 & 53.02 & 73.47 & 62.86 & 67.93 & 69.33 & 67.80 & 71.70 & 69.61 \\
\bottomrule
\end{tabular}
}

\caption{Ablation studies on data composition and quality. We report Pass@1 accuracy (\%) across mathematical reasoning and general multimodal benchmarks. All experiments used MiMo-VL-7B-SFT-2508 as the base model.}
\label{tab:ablation}
\end{table*}

DeepVision spans two data domains—multimodal math and visual logic, which differ in reasoning paradigms. Multimodal math requires extracting visual evidence and applying mathematical knowledge (e.g., formulas, theorems, computations) to reach an answer. In contrast, visual logic is driven mainly by visual cues (e.g.,object positions, spatial relations, and patterns), with little reliance on explicit mathematical knowledge.
\citet{visiong1} points out that mixing heterogeneous domains may introduce interference and conflicting gradients, potentially harming learning. This motivated us to examine whether introducing visual-logic data is indeed beneficial, and how each domain contributes to the final performance.

We performed controlled ablations by varying the training data composition while keeping the data exposure comparable. In our full setting (DeepVision-103K$_{200}$), our final model, MiMo-VL-7B-DeepVision, was trained for 200 steps on a 3:1 mixture of multimodal math (77K) and visual logic (26K). We evaluated three single-domain counterparts:

\begin{itemize}
    \item \textbf{Math-77K$_{150}$}: math only for 150 steps (same math exposure as DeepVision$_{200}$).
    \item \textbf{Math-77K$_{200}$}: math only for 200 steps (same total exposure as DeepVision$_{200}$).
    \item \textbf{Visual-logic-26K$_{50}$}: visual logic only for 50 steps (same visual logic exposure as DeepVision$_{200}$).
\end{itemize}

Results in Table~\ref{tab:ablation} show that scaling math training is consistently beneficial: both math-only variants outperform the base model, and extending training from 150 to 200 steps improves every benchmark. Howerver, math alone is not sufficient to reach the best performance. Under the same total exposure, Math-77K$_{200}$ underperforms the mixed setting on math average (69.02\% vs. 70.10\%) with a clear gap on LogicVista (63.53\% vs. 65.92\%). 

These results indicate that introducing visual logic data is valuable, and is further supported by the visual logic-only setting (Visual-logic-26K$_{50}$), which improves over the foundation model across all benchmarks, demonstrating positive transfer from visual logic to both mathematical and general evaluations. We attribute these gains to two factors: (i) spatial reasoning and pattern recognition are broadly useful primitives shared across mathematical and general multimodal tasks, and (ii) visual logic training directly strengthens these primitives while multimodal math alone does not sufficiently cultivate them.

\subsection{Necessity of query correctness verification.}

After pass-rate filtering, we obtained 99k samples calibrated to the model's capability. To ensure the validity of the reward signals in RLVR, we further applied Gemini-3.0-Flash to remove samples with garbled text or image–text mismatches, and  filtered out samples whose answers were inconsistent with Gemini’s solutions, discarding an additional 22K samples. However, \citet{rom,spuriousreward} have suggested that LLMs can improve even under spurious rewards, raising doubts about whether strict query correctness is essential for RLVR. To investigate this, we evaluated an unverified variant (Unverified-125K$_{200}$) which was trained 200 steps on the 99k unverified math data and 26k visual logic data.

Table~\ref{tab:ablation} shows that Unverified$_{200}$ improves over the base model, but remains substantially worse than DeepVision$_{200}$ (67.93\% vs. 70.10\% on math average; 69.61\% vs. 70.92\% on general average). This indicates that query correctness verification is necessary because corrupted inputs or incorrect answers hinder the model’s progress，highlighting that accurate and reliable reward signals are crucial for multimodal RLVR.

\section{Conclusion}
We present \textbf{DeepVision-103K}, a large-scale and verifiable multimodal dataset for RLVR, curated from diverse real-world K12 sources via a three-stage pipeline of validity filtering, pass-rate-based difficulty calibration, and query correctness verification. DeepVision-103K incorporates wide-ranging multimodal mathematical problems and visual logic problems, and covers major visual categories including geometry, analytic plots, charts, and real-world items in mathematical contexts. Training on DeepVision-103K yields top performance on both mathematical and general multimodal tasks. Our further analysis reveals enhanced visual perception, reflection and reasoning capabilities for models trained on DeepVision-103K. We point out multimodal math data and visual logic data contribute to each other in multimodal reasoning, and show the importance of query correctness in multimodal RLVR training.

\section{Limitations}

While DeepVision-103K substantially increases visual diversity, the distribution is imbalanced (e.g., planar geometry dominates), and some rare element types remain underrepresented. our pipeline relies on strong external models (e.g., Gemini) for query correctness verification, which may introduces potential bias and additional cost, and may filter out a small portion of valid but hard samples. Our dataset focuses on K12-level problems with unique final answers to enable verifiable rewards; thus it does not fully cover open-ended mathematical tasks (e.g., proof writing, multi-solution problems) that require richer evaluation signals.

\bibliography{custom}

\appendix

\section{Visual Examples}
\label{appendix:visual_examples}
In this section, we present cross-category visual combination examples in DeepVision-103k.

\begin{figure}[!h]
    \centering
    \includegraphics[width=0.48\textwidth]{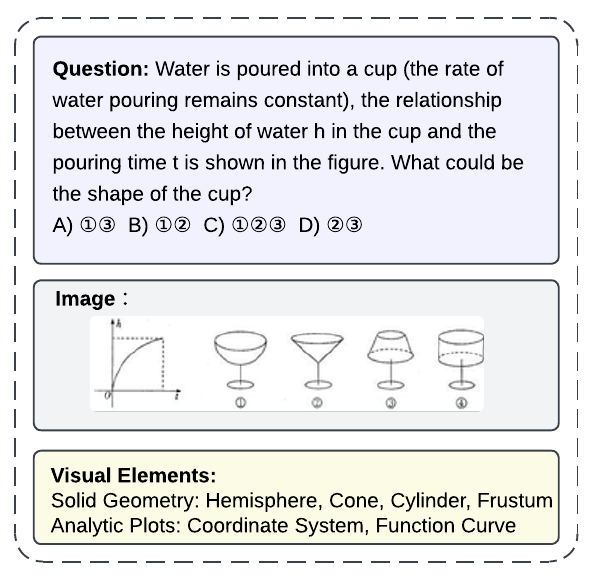}
    \caption{Solid Geometry \& Analytic Plots.}
    \label{fig:sa}
\end{figure}
\vspace{-2mm}
\begin{figure}[!h]
    \centering
    \includegraphics[width=0.48\textwidth]{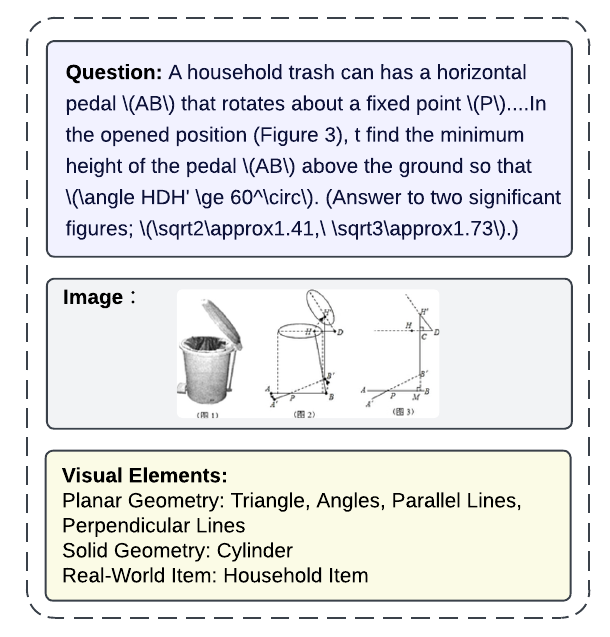}
    \caption{Planar Geometry \& Solid Geometry \& Real-World Item.}
    \label{fig:psr}
\end{figure}

\section{Visual Elements Annotation}
\label{visual-anno}

To characterize the distribution of visual elements in DeepVision-103K and existing datasets, we constructed a visual annotation taxonomy based on~\cite{a,b}. We then instructed GPT-5 mini to annotate visual elements in each dataset according to the proposed taxonomy (Table \ref{tab:visual-taxonomy}). We set the decoding temperature to 0.1 and the thinking budget to \texttt{low}.

\begin{table*}[t]
\centering
\small
\setlength{\tabcolsep}{6pt}
\renewcommand{\arraystretch}{1.2}
\begin{tabular}{p{0.17\textwidth} p{0.78\textwidth}}
\toprule
\textbf{Category} & \textbf{Fine-grained types} \\
\midrule
\texttt{planar\_geometry} &
Right Triangle; Equilateral Triangle; Triangle; Square; Rectangle; Rhombus; Parallelogram; Trapezoid; Quadrilateral; Circle; Semicircle; Sector; Arc; Parallel Lines; Perpendicular Lines; Tangent; Chord; Angle; Right Angle \\
\texttt{solid\_geometry} &
Cube; Cuboid; Prism; Pyramid; Tetrahedron; Sphere; Cylinder; Cone; Frustum; Hemisphere; Net; Orthographic View \\
\texttt{analytic\_plot} &
Linear Graph; Parabola; Hyperbola; Sinusoidal Curve; Exponential Curve; Analytic Circle; General Function Curve; Coordinate System; Number Line; Scatter Points; Inequality Region; Equation \\
\texttt{data\_chart} &
Table; Bar Chart; Line Chart; Pie Chart; Donut Chart; Histogram; Box Plot; Stem-and-Leaf Plot \\
\texttt{schematic\_diagram} &
Flowchart; Circuit; Force Diagram; Tree Diagram; Venn Diagram; Linear Arrangement \\
\texttt{real-world item} &
Character; Plant; Scientific Tool; Vehicle; Architecture; Household Item; Apparel; Food; Real Object; Scene; Map \\
\bottomrule
\end{tabular}
\caption{Visual-element annotation taxonomy used in this work.}
\label{tab:visual-taxonomy}
\end{table*}

\section{Data Construction}
\label{appendix:data cons}

\subsection{Difficulty Filtering}
\label{appendix:Difficulty Filtering}
For the math subset, we retain all examples whose pass rate falls in $[\tfrac{1}{8}, \tfrac{4}{8}]$. For the easier range $[\tfrac{5}{8}, \tfrac{7}{8}]$, we do not include all available data due to its large volume; instead, we selectively sample from this range by prioritizing knowledge points that are under-represented in the $[\tfrac{1}{8}, \tfrac{4}{8}]$ portion, thereby improving coverage while keeping the dataset size manageable.

\begin{figure}[!h]
    \centering
    \includegraphics[width=0.48\textwidth]{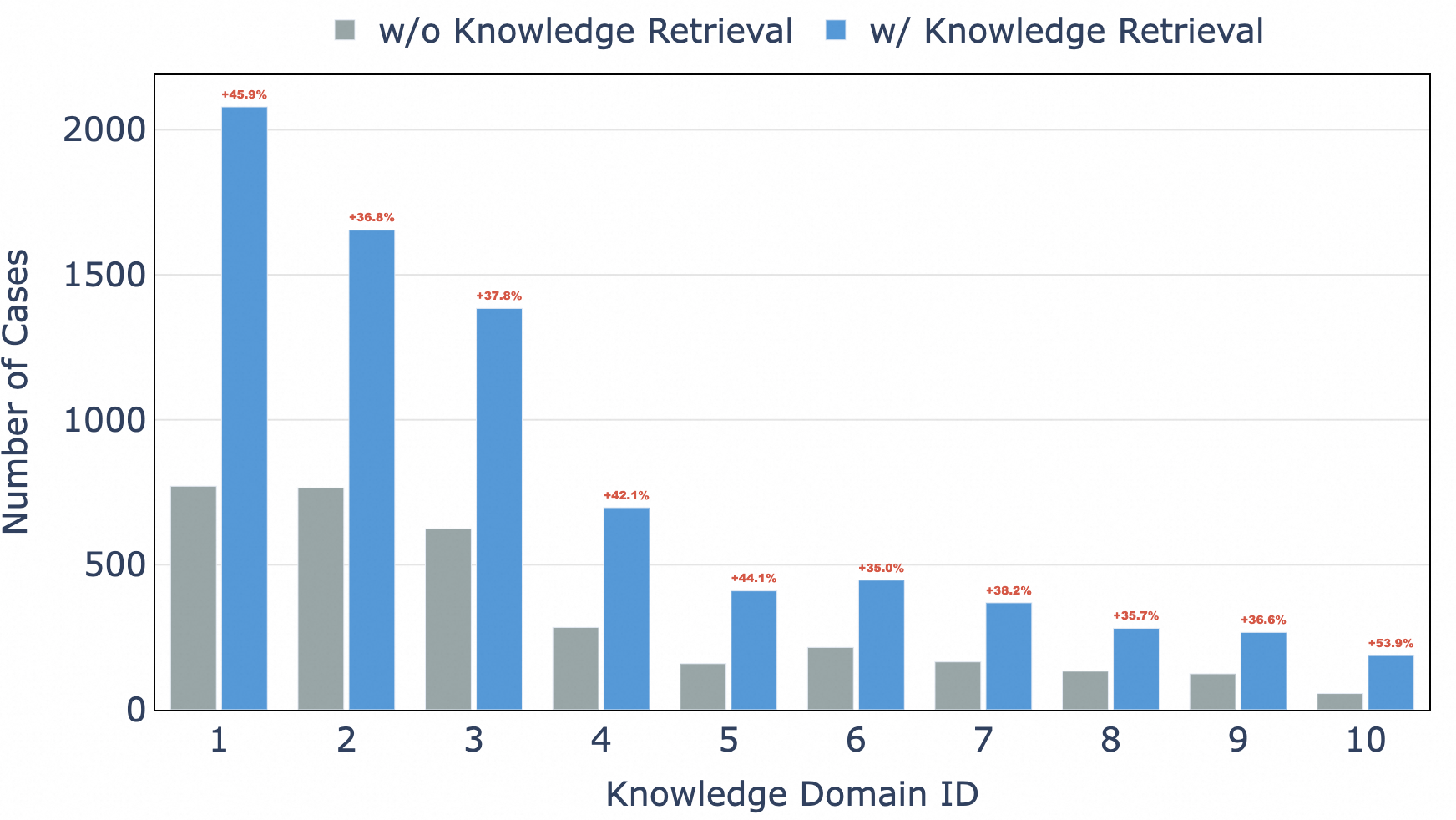}
    \caption{Top 10 Knowledge-based retrieval. The x-axis IDs correspond to knowledge domains listed in Table~\ref{tab:knowledge_retrieval}.}
    \label{fig:recall}
\end{figure}

Our empirical study reveals that training with knowledge-guided retrieved data outperforms training without retrieval under the same training steps. We present the top 10 retrieval knowledge points in Figure~\ref{fig:recall} and Table~\ref{tab:knowledge_retrieval}. 

\begin{table*}[h]
\centering
\resizebox{\textwidth}{!}{%
\begin{tabular}{clcccc}
\toprule
\textbf{ID} & \textbf{Knowledge Domain} & \textbf{w/o Retrieval} & \textbf{w/ Retrieval} & \textbf{Increase}  \\
\midrule
1 & Circle $\rightarrow$ Inscribed and Circumscribed & 771 & 2,079 & +1,308  \\
2 & Triangle $\rightarrow$ Angle of Elevation and Depression & 765 & 1,654 & +889  \\
3 & Circle $\rightarrow$ Tangency & 624 & 1,384 & +760  \\
4 & Circle $\rightarrow$ Perpendicular Chord Theorem & 284 & 697 & +413  \\
5 & Conic Sections $\rightarrow$ Hyperbola & 159 & 410 & +251  \\
6 & Figure Relationships $\rightarrow$ Inscribed and Circumscribed & 215 & 447 & +232  \\
7 & Spatial Relationships $\rightarrow$ Parallelism \& Perpendicularity & 165 & 369 & +204  \\
8 & Conic Sections $\rightarrow$ Parabola & 133 & 281 & +148  \\
9 & Triangle $\rightarrow$ Criteria for Similar Triangles & 124 & 267 & +143  \\
10 & Spatial Relationships $\rightarrow$ Angle between Line \& Plane & 56 & 187 & +131  \\
\bottomrule
\end{tabular}%
}
\caption{Top 10 Knowledge Domains by Retrieval Gap}
\label{tab:knowledge_retrieval}
\end{table*}

For visual logic data, we used tetris, maze, chess data from Zebra-CoT \cite{zebracot} and game data from GameQA-140K \cite{gamerl} with pass rate at $[\tfrac{3}{8}, \tfrac{4}{8}]$. This choice was made to broaden the training data distribution while keeping the dataset size manageable.

\subsection{Correctness Verification}
\label{appendix:Correctness Verification}
To ensure data reliability, we used \textsc{Gemini 3 Flash} as an automated verifier. For each instance, it jointly inspected the input image, question text, reference answer, and outputs a \emph{label} with a \emph{judge trace}. The verifier follows a deterministic decision rule with a strict precedence hierarchy (Table \ref{tab:correctness_verification_labels}). 
\begin{itemize}
\item \textbf{Input Corectness.} The verifier first checks data integrity and rejects the instance if any of the following labels is triggered:
  \texttt{ERR\_IMG\_MISSING}, \texttt{ERR\_TEXT\_MISSING}, or \texttt{ERR\_MISMATCH}.
  \item \textbf{Answer Correctness.} For well-formed inputs, the verifier evaluates the reference answer; if incorrect, it outputs \texttt{CORRECTION} with a revised solution.
  \item \textbf{Acceptance.} An instance is marked as correct only when no input-level or answer-level errors are detected.
\end{itemize}

We discarded all instances flagged as \texttt{CORRECTION} rather than replacing the answer, to avoid introducing noise from automatic edits.

\begin{table*}[!h]
  \centering
  \small
  \setlength{\tabcolsep}{8pt}
  \renewcommand{\arraystretch}{1.25}
  \begin{tabular}{p{0.28\linewidth} p{0.22\linewidth} p{0.44\linewidth}}
    \toprule
    \textbf{Label} & \textbf{Category} & \textbf{Trigger} \\
    \midrule
    \texttt{ERR\_IMG\_MISSING} &
    Image quality issue &
    Image is missing, unreadable, or lacks essential visual information. \\
    \texttt{ERR\_TEXT\_MISSING} &
    Missing text &
    Question text misses key conditions/values, making the task unsolvable. \\
    \texttt{ERR\_MISMATCH} &
    Image--text mismatch &
    Image content conflicts with the question statement. \\
    \texttt{CORRECTION} &
    Incorrect reference answer &
    Data are valid, but the reference answer is incorrect; return the corrected solution/answer in \LaTeX. \\
    \texttt{1} &
    Perfect match &
    Image/text are complete and consistent, and the reference answer is correct. \\
    \bottomrule
  \end{tabular}
  \caption{Verification labels used by \textsc{Gemini 3 Flash}. Exactly one label is returned per instance.}
  \label{tab:correctness_verification_labels}
\end{table*}

\subsection{Data Licenses}
We list the data collection protocol of our data sources in Table \ref{tab:data_licenses}.

\begin{table*}[!h]
\centering
\small
\setlength{\tabcolsep}{8pt}
\renewcommand{\arraystretch}{1.15}
\begin{tabularx}{\textwidth}{@{} l l X @{}}
\toprule
\textbf{Data Source} & \textbf{License} & \textbf{URL} \\
\midrule
MM-MathInstruct-3M~\citep{mmmath} & Apache 2.0 &
\url{https://huggingface.co/datasets/MathLLMs/MM-MathInstruct} \\
MultiMath-300K~\citep{multimath} & Unset &
\url{https://huggingface.co/datasets/pengshuai-rin/multimath-300k} \\
Zebra-CoT~\citep{zebracot} & CC BY-NC 4.0 &
\url{https://huggingface.co/datasets/multimodal-reasoning-lab/Zebra-CoT} \\
GameQA-140K\cite{gamerl} & MIT & \url{https://huggingface.co/datasets/Code2Logic/GameQA-140K}\\
PuzzleVQA\cite{puzzle} & Unset & \url{https://huggingface.co/datasets/declare-lab/PuzzleVQA}\\
\bottomrule
\end{tabularx}
\caption{Licenses and usage permissions for the data sources used in this work.}
\label{tab:data_licenses}
\end{table*}


\section{Training Details}
\label{app:training setting}
We used \texttt{verl} as the training framework. Configurations for training DeepVision series models are listed in Table~\ref{tab:hyperparameters}.
\begin{table}[h]
\centering
\begin{tabular}{lc}
\toprule
Config & Value \\
\midrule
lr & 1e-6 \\
kl\_coef & 1e-3 \\
max\_prompt\_length & 2K \\
max\_response\_length & 16K \\
gen\_batch\_size & 512 \\
train\_batch\_size & 256 \\
mini\_batch\_size & 64 \\
micro\_batch\_size & 32 \\
group\_filtering & acc \\
clip\_ratio\_low & 1e-3 \\
clip\_ratio\_high & 1e-4 \\
temperature & 1.0 \\
rollout.n & 16 \\
total\_training\_steps & 200 \\
\bottomrule
\end{tabular}
\caption{Configurations for training DeepVision series models.}
\label{tab:hyperparameters}
\end{table}

We used 32 H20 GPU for a single training, a training step cost 0.5h. We used the following prompt template during training and evaluation.

\begin{prompt}{Training / Evaluation Prompt Template}
You are a multimodal reasoning assistant. You receive images and texts, perform step-by-step reasoning (including re-checking the image) before producing the final answer. Please provide a clear, concise answer inside \textbackslash boxed\{\} tag. For multiple choice questions, put only the letter like \textbackslash boxed\{A\} without any additional text. For fill-in-the-blank and problem-solving questions, put only the final answer.
\end{prompt}

\section{Evaluation Details}
\label{app:eval setting}

We provide detailed information about the benchmarks used for evaluation and the inference hyperparameters for each model.

\subsection{Benchmarks}

We evaluated our models across three categories of benchmarks, as summarized in Table~\ref{tab:benchmarks}.

\begin{table}[h]
\centering
\resizebox{\columnwidth}{!}{
\begin{tabular}{@{}llcc@{}}
\toprule
\textbf{Category} & \textbf{Benchmark} & \textbf{\#Samples} & \textbf{Reference} \\
\midrule
\multirow{4}{*}{Multimodal Math} 
& WeMath & 1,740 & \citep{bench:wemath} \\
& MathVision & 3,040 & \citep{bench:mathvision} \\
& $\text{MathVerse}_{\text{vision}}$ & 788 & \citep{bench:mathverse} \\
& LogicVista & 448 & \citep{bench:logicvsita} \\
\midrule
\multirow{3}{*}{General Multimodal} 
& $\text{M}^{3}\text{CoT}$ & 2,318 & \citep{m3cot} \\
& $\text{MMMU}_{\text{val}}$ & 900 & \citep{bench:mmmu} \\
& $\text{MMMU}_{\text{Pro\_full}}$ & 1,730 & \citep{mmmu_pro} \\
\midrule
\multirow{2}{*}{Text-only Math} 
& AIME 2025 & 30 & \citep{bench:aime25} \\
& HMMT 2025 & 30 & \citep{bench:hmmt25} \\
\bottomrule
\end{tabular}
}
\caption{Overview of evaluation benchmarks.}
\label{tab:benchmarks}
\end{table}

\subsection{Inference Hyperparameters}

We used different inference hyperparameters for different model families to ensure optimal performance. The detailed configurations are listed in Table~\ref{tab:inference_params}.

\begin{table}[h]
\centering
\resizebox{\columnwidth}{!}{
\begin{tabular}{@{}lccc@{}}
\toprule
\textbf{Parameter} & \textbf{Qwen3-VL-Thinking} & \textbf{Qwen3-VL-Instruct} & \textbf{MiMo-VL-(SFT/RL)} \\
\midrule
top\_p & 0.95 & 0.8 & 0.95 \\
top\_k & 20 & 20 & -- \\
temperature & 1.0 & 0.7 & 0.3 \\
repetition\_penalty & 1.0 & 1.0 & -- \\
presence\_penalty & 0.0 & 1.5 & -- \\
max\_tokens & 32,768 & 32,768 & 32,768 \\
\bottomrule
\end{tabular}
}
\caption{Inference hyperparameters for each model family.}
\label{tab:inference_params}
\end{table}

For Qwen3-VL-DeepVision models, we adopted the same hyperparameters as Qwen3-VL-Instruct. For MiMo-VL-DeepVision, we adopted the same hyperparameters as MiMo-VL.

\subsection{Evaluation Method}

For each benchmark, we first calculated accuracy with MathVerify~\cite{mathverify}, then prompted GPT-5-mini to re-judge cases marked as incorrect by MathVerify to reduce false negatives caused by parsing errors, equivalent expressions, or formatting variations. We used the revised judgment as the final label.

\section{Training Curves}
\label{training curves}
\begin{figure}[!h]
    \centering
    \includegraphics[width=0.48\textwidth]{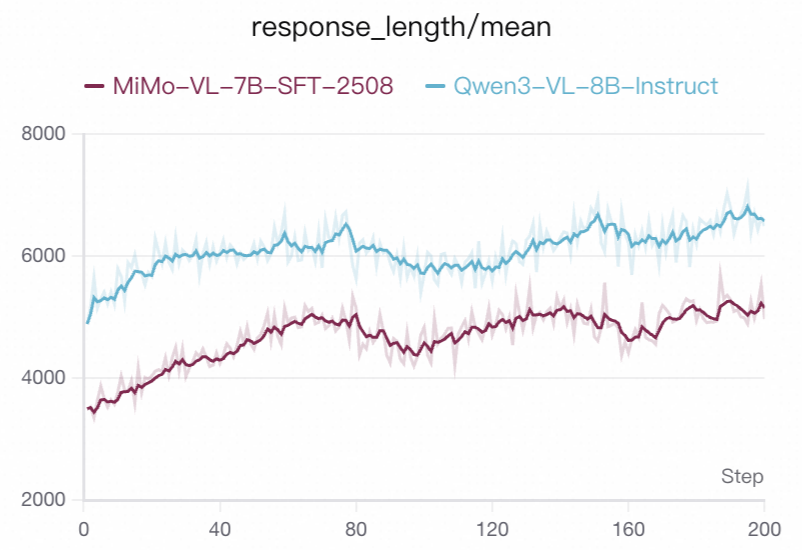}
    \caption{Increasing response length.}
    \label{fig:resp}
\end{figure}

\begin{figure}[!h]
    \centering
    \includegraphics[width=0.48\textwidth]{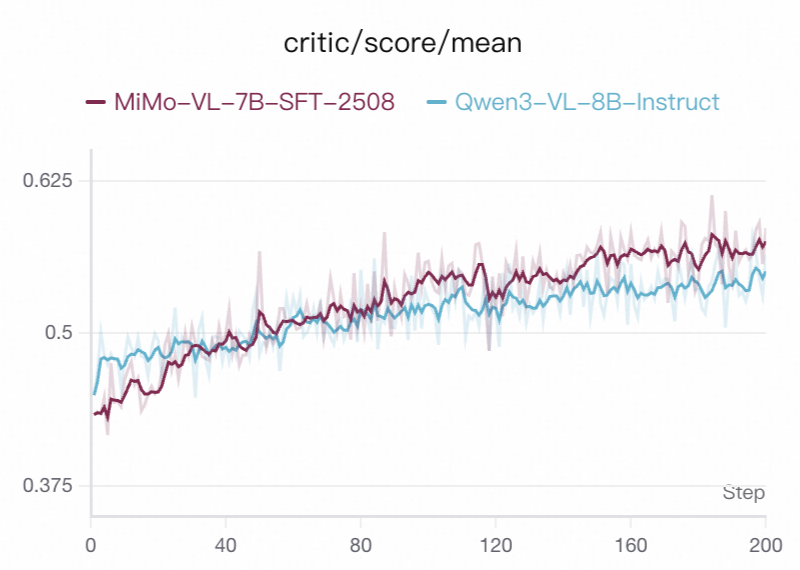}
    \caption{Upward rewards.}
    \label{fig:critic}
\end{figure}

\begin{figure}[!h]
    \centering
    \includegraphics[width=0.48\textwidth]{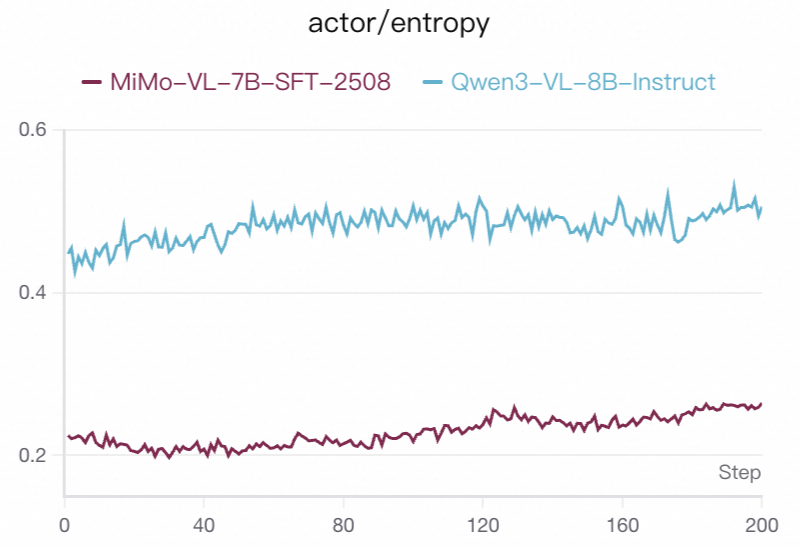}
    \caption{Stable entropy.}
    \label{fig:entropy}
\end{figure}

This section presents the training dynamics on DeepVision-103K, including response length (Figure \ref{fig:resp}), trainset rewards (Figure \ref{fig:critic}) and entropy (Figure \ref{fig:entropy}).

\section{Potential Risks}
We do not anticipate significant potential risks from this work. DeepVision-103K is derived from publicly available K12-level educational content and is designed for verifiable-answer multimodal reasoning rather than sensitive decision-making. The dataset contains no personal identifiers, and our curation process filters out corrupted or unsafe samples.

\end{document}